# Autonomous UAV Navigation for Search and Rescue Missions Using Computer Vision and Convolutional Neural Networks

Luka Šiktar[1], Branimir Ćaran[1], Bojan Šekoranja[1] and Marko Švaco[1]

[1] Faculty of Mechanical Engineering and Naval Architecture, University of Zagreb, Croatia
`luka.siktar@fsb.hr, https://fsb.unizg.hr`

**Abstract.** In this paper, we present a subsystem, using Unmanned Aerial Vehicles (UAV), for search and rescue missions, focusing on people detection, face recognition and tracking of identified individuals. The proposed solution integrates a UAV with ROS2 framework, that utilizes multiple convolutional neural networks (CNN) for search missions. System identification and PD controller deployment are performed for autonomous UAV navigation. The ROS2 environment utilizes the YOLOv11 and YOLOv11-pose CNNs for tracking purposes, and the dlib library's CNN for face recognition. The system detects a specific individual, performs face recognition and starts tracking. If the individual is not yet known, the UAV operator can manually locate the person, save their facial image and immediately initiate the tracking process. The tracking process relies on specific keypoints identified on the human body using the YOLOv11-pose CNN model. These keypoints are used to track a specific individual and maintain a safe distance. To enhance accurate tracking, system identification is performed, based on measurement data from the UAV's IMU. The identified system parameters are used to design PD controllers that utilize YOLOv11-pose to estimate the distance between the UAV's camera and the identified individual. The initial experiments, conducted on 14 known individuals, demonstrated that the proposed subsystem can be successfully used in real time. The next step involves implementing the system on a large experimental UAV for field use and integrating autonomous navigation with GPS-guided control for rescue operations planning.

**Keywords:** UAV, Autonomous System, Search and Rescue, Computer Vision, Convolutional Neural Network

## 1    Introduction

Search and rescue operations are integral to effective security management. These rapid and high intensity missions often require rescue specialists to risk their lives [1]. Emerging technologies, such as Unmanned Aerial Vehicles (UAV) equipped with detection and tracking capabilities based on computer vision, can significantly assist specialists in various scenarios [2], [3]. Currently, most research papers focus on deep learning methods for people detection, face recognition and tracking using UAVs. Sambolek and Ivasić-Kos proposed a people detection method using fine-tuned YOLOv4 model, trained on custom dataset, evaluated under various weather conditions [4]. Melkumyan and Mkrtchyan evaluated Local Binary Pattern Histogram, FaceNet, and Face_recognition algorithm for face detection and recognition using UAVs [5]. Priambodo et al. utilized DJI Tello UAV for face detection and tracking using Haar cascade bounding boxes, with a PID controller designed to maintain a relative distance to person's face [6]. Mercado-Ravell et al. proposed face detection and face tracking algorithm that calculates the circular area around the face. The proposed UAV control is performed using Kalman Filter [7]. Ollachica et al. proposed highly accurate and fast face detection and recognition deep learning method, which requires fine-tuning for every individual [8]. Shen et al. proposed a system using YOLOv3 for people detection, the LLC method for recognition, and five facial keypoints for face tracking. PD controller is also proposed for UAV tracking [9]. Hakani and Rawat proposed real-time UAV people detection using YOLOv9, performed on NVIDIA Jetson Nano onboard computer [10].

We propose a UAV-based search subsystem, dedicated to search and rescue opearions, that incorporates people detection, face recognition and body keypoint tracking using DJI Tello UAV. The research serves as the proof of concept for autonomous search and rescue UAV system. People detection is performed using YOLOv11, filtered to detect only people. Face extraction and recognition is performed using dlib's face recognition deep learning model [11]. Body keypoint extraction, used for relative UAV-person distance approximation, is achieved using YOLOv11-pose model. The drone tracks specified individuals by using distance approximation as the input to three separate PD controllers for forward-backward, up-down and yaw rotation movements. PD controllers regulate relative distance and UAV's rotation and altitude, based on individual's position in the UAV camera's image. The proposed solution, developed for ROS2 environment, was evaluated for real-time indoor and outdoor usage, achieving results comparable to traditional computer vision approaches. The main contributions of this research are the integration of a compact deep learning-based solution, utilizing YOLOv11, face recognition CNN and YOLOv11-pose to detect, recognize and track the individuals at a relative distance using monocular UAV's camera. Another contribution is the use of UAV-specific data-driven system identification and PD controller tuning to enhance tracking performance.

## 2 Materials

In this paper, the DJI Tello UAV is utilized. The Tello is an educational indoor UAV equipped with 720p video camera, Intel 14-core processor, an additional downward-facing camera and barometer for altitude hold. The UAV and UAV coordinate system is shown in Fig.1. left, also UAV video camera footage with displayed mode data and crosshair in the center are shown in Fig 1., right. The Tello platform is preferred for testing due to its connectivity and compatibility with ROS/ROS2. People detection, face recognition and body tracking are achieved by transmitting the video stream from UAV to the Jetson NX Xavier single board computer via Wi-Fi. The Jetson NX Xavier has 384-core NVIDIA Volta GPU with 48 Tensor Cores, 6-core NVIDIA Arm 64-bit

CPU, and 8GB of RAM. After processing the video frames, velocity commands are transmitted back to the UAV through ROS2 framework.

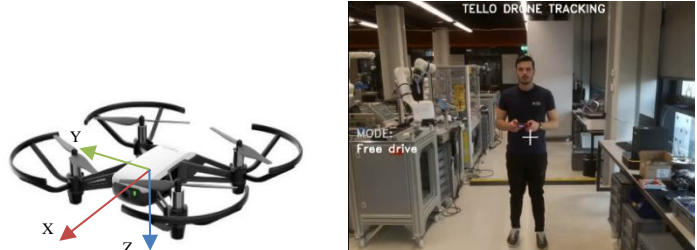

**Fig.1. Left.**Tello UAV [12], **Right.** Image from UAV camera

To identify people and their locations on video stream, the YOLOv11 CNN is utilized. YOLOv11 is a state-of-the-art object detection model from Ultralytics, pretrained on the COCO dataset. The YOLOv11s model is specifically chosen due to its performance and size, enabling real-time detection on 15 Hz. Face recognition is performed using dlib's face recognition deep learning model [11]. The body tracking process utilizes the YOLOv11-pose to identify the locations of specific body keypoints from the video stream. Shoulder and hip keypoints are used to approximate the relative distance between the UAV and the person. Accurate tracking is further improved through data-driven system identification and controller tuning, conducted using Matlab's tools.

## 3 Methodology

### 3.1 System identification and PD controller tuning

The initial step of the proposed solution involves system identification and tuning of three individual PD controllers for the UAV. The data-driven system identification includes recording the telemetry data from UAV's Inertial Measurement Unit (IMU) and identifying the dynamic model. By applying specific control inputs that correspond to distinct actions, linear forward-backward movement, vertical up-down movement, and yaw rotation, three separate PD controllers are designed. Successful system identification requires exciting all system's poles using specifically designed input signals, in this case, command velocities and throttle applied to the UAV. Due to the slowness and inertia of the system, it was assumed that the UAV dynamics could be described by linear differential equations to simplify the model identification process based on input values of velocities in the X and Z directions, as well as rotation around the Z-axis.

For each specified action, the predefined input signal patterns are performed, and the UAV's output velocities are measured, as shown in Fig 2 graph1 and graph2. Input and output data are monitored and logged to identify the dynamic models for each distinct action using Matlab's System Identification Toolbox, which enables accurate modeling of the UAV's behavior. Controller tuning is performed by modelling the system and performing analysis using Simulink, as shown for X-axis in Fig.2. P, PD and PID controllers are experimentally evaluated to identify the suitable option for smooth operations with minimal overshot and no oscillations. PD controllers are selected for all three actions due to their responsiveness and stability tradeoff. The roll/X-axis PD controller reduces the error of calculated person-UAV distance, while for the yaw/Z-axis linear movement and yaw angular rotation, independent PD controllers reduce the distance between the center of the image, i.e. the center of the image, marked with crosshair in Fig. 1 and the shoulder's midpoint, detected using YOLOv11-pose, shown in Fig 3 (f).

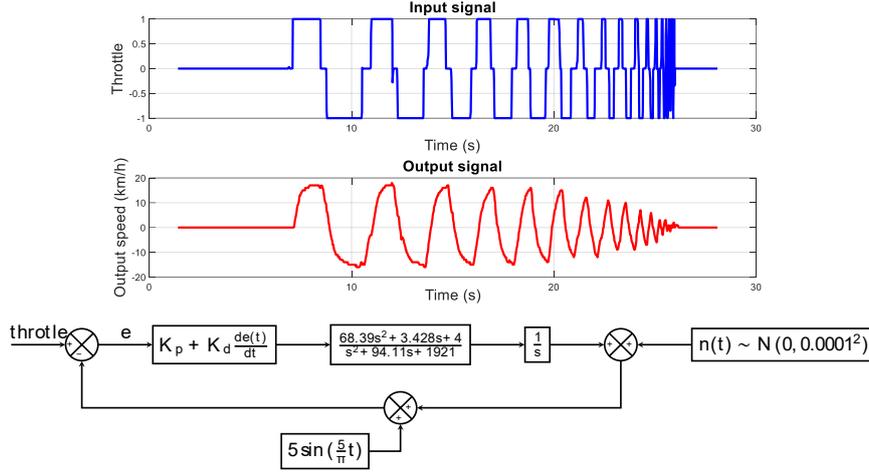

**Fig.2. Upper gaph1**: Input throttle signal recorded for the X-axis system identification, **Lower graph2**: Output velocities of a UAV for the X-axis, used for system identification, **bottom**: Proposed system model for X-axis, obtained by system identification. The system is controlled with PD controller, with added white noise and sine disturbance that simulates approximated UAV-person distance changes over time to evaluate the system performance.

### 3.2 People detection

For people detection, You Only Look Once 11 (YOLOv11) CNN model is used. YOLOv11 is a real-time object detection model pretrained on the COCO dataset. It is a state-of-the-art model in terms of feature extraction and processing speed [13]. YOLOv11 achieves a higher mean Average Precision (mAP) score on the COCO dataset compared to the predecessors while being more computationally efficient. For this task, "Detect" mode of YOLOv11 is used. The "Detect" mode accurately predicts the classes and precise locations of 80 labeled classes within the COCO dataset. To focus on people detection, the output predictions are filtered to retain only those detections corresponding to the "person" class. To enable tracking through consecutive frames, it is necessary to identify every individual and assign them a specific ID. This is achieved by utilizing YOLOv11's built-in tracking mode, which enables connections between detections across frames. The results of the people detection process on images captured by the UAV are illustrated in Fig. 3(a) and (b).

### 3.3 Face detection and recognition

Following the people detection process, face recognition is performed to determine if the searched individual is present on the image. The face recognition is performed using dlib's deep learning algorithm based on ResNet architecture. The process begins by extracting the detected individuals using bounding boxes provided by YOLOv11 tracking mode. Face detection, based on Histogram of Oriented Gradients (HOG), is performed to extract the face from the individual's bounding box. The detected face is aligned to a standard position required for accurate face recognition. From the aligned face, key features are extracted using a deep learning model to generate the 128-dimensional feature vector. The vector is then compared with the vector generated from the searched individual's face image, by calculating Euclidean distance. If the distance is below the standard threshold 0.6, used in [11], the searched individual is recognized. The process of face detection and extraction is shown in image Fig.3. (c) and (d). Once the searched individual is recognized, their ID and bounding box is stored and tracked for further analysis.

### 3.4 Body tracking

Tracking of a specified person is accomplished using pose estimation information provided by the YOLOv11-pose CNN model. YOLOv11-pose identifies 17 body keypoints including the eyes, ears, shoulders, elbows and hips. Unlike recent research [5-9] which focus on maintaining a constant relative distance between the UAV's camera and the person's face, our algorithm maintains a constant relative distance between the UAV's camera and the individual's body, using distance between person's shoulders and hips. The motivation for selecting body keypoint approach comes from the need to reliably follow a person in scenarios where their face is not visible. Additionally, the distance between the shoulders and hips is the reference dimension because it remains invariant to the person's orientation relative to the camera. A potential limitation of the system is the situation when a person leans toward the camera, distorting the shoulder to hip distance on 2D camera image. This issue is addressed by restricting the UAV's motion when rapid changes in the calculated distance are detected. Such rapid changes can be identified as posture variations rather than changes in physical proximity.

The shoulder-hip distance (Fig.3 (f), white line), measured in pixels, is used to calculate the approximate relative distance between the average European male height of 180 cm, or female height of 171 cm, and the camera using the formula (1).

$$y = k_1 x^2 + k_2 x + k_3 \ \ [\text{cm}] \tag{1}$$

$$y - \text{relative UAV} - \text{person distance [cm]}$$
$$x - \text{distance between center of hips and center of shoulders [px]}$$

The parameters in (1) were determined experimentally by deriving the polynomial equation for measured distances, accurately calibrated in the range 0 to 6 meters.

### 3.5 Complete application

The application is built using the ROS2 framework, integrating DJI Tello driver, Joy package for manual control, and search and rescue subsystem. The created application works as shown in Fig.3 (g). PS4 joystick is used for manual UAV control ("Free mode"). In this mode, the UAV can be maneuvered without utilizing the proposed search and rescue subsystem. If an image of the searched individual exists, the 128-dimensional template vector embedding is calculated and stored for face recognition, using dlib's model. When the search process is initiated using joystick button ("Search mode"), the UAV will start rotating and scanning the environment for the specified individual. Upon recognizing the individual, the UAV centers their body within the camera's image frame and waits for the operator to initiate the tracking process. The tracking is initiated using joystick button ("Follow/Track mode") and UAV tracks the detected body while maintaining the specified distance, set to 2 meters in this case.

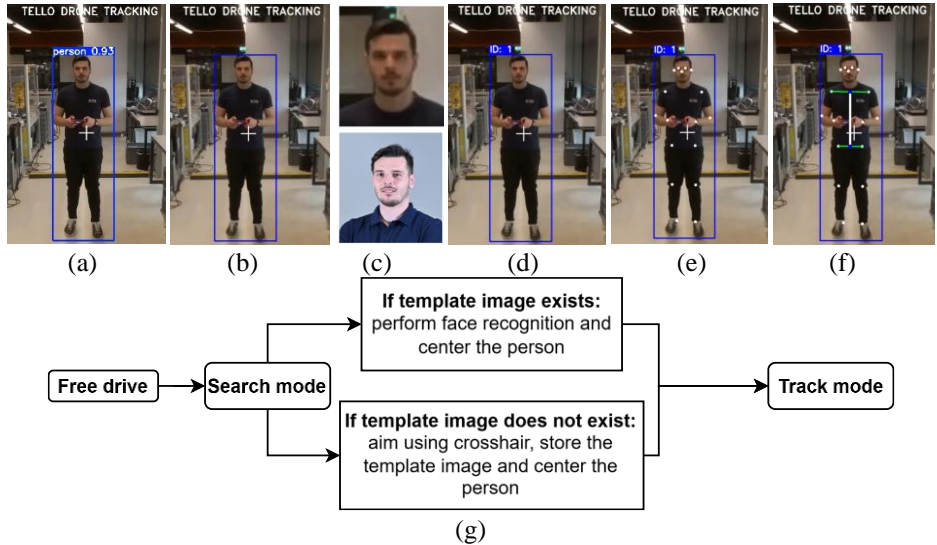

**Fig.3.** (a) People detection, standard YOLOv11, (b) Filtered YOLOv11, people detection results, (c) Extracted face image (top) and template image (bottom), (d) Face recognition and ID assignment, (e) YOLOv11-pose body keypoints results, (f) White line presents constructed shoulder to hip distance used for distance approximation, (g) Complete system workflow

In case there is no existing image of the searched individual, the UAV can be manually operated to align its camera with the specified individual using crosshair displayed in the center of the image. Once the person is centered, activating "Search Mode" saves their template image for subsequent recognition. From this point, the remaining processes proceed as described. The complete application is performed using ROS2 environment on Jetson NX Xavier, by acquiring the video stream from the UAV via Wi-Fi, using PS4 joystick controller for basic control commands, connected via Bluetooth.

The proposed system calculates UAV command velocities in X and Z axes and yaw velocity. The velocities are sent to the UAV using ROS2 via Wi-Fi.

## 4       Experiments

The conducted experiments aimed to evaluate the performance of the proposed system on low quality UAV's camera. Despite the quality of the 720p video stream, all the proposed deep learning methods demonstrated excellent performance. YOLOv11 and YOLOv11-pose models efficiently detected all specified individuals and their body keypoints for tracking without any significant issues. The face recognition model performed satisfactorily even though the UAV's camera feed is lower resolution than template images so the algorithm can cause false positive (FP) and false negative results (FN). The proposed deep learning models' limitations are the same as UAV's flight limitations, which prevent flying in dark and highly illuminated environments.

The challenge encountered during testing was data transmission between the UAV and the Jetson NX Xavier via Wi-Fi. The wireless connection between the UAV's camera and the computer led to higher risk of image loss, leading to occasional glitch frames. The glitch frames could cause losing traction and failing to maintain tracking procedure. If the algorithm fails while tracking or searching, the automatic re-conformation procedure would start. The re-conformation procedure consists of repeating the face-recognition process using the stored template images.

The system's real-time performance was also evaluated by analyzing the operational speeds of deep learning models and ROS2 framework. The UAV video stream is obtained at approximately 30 Hz, with the YOLOv11 model processing frames and 30 Hz and the YOLOv11-pose model at 15Hz. Face recognition, that was initiated only in search phase, operated at 5 Hz, so the yaw velocity of the UAV during search is not recommended to exceed 45º/s. Velocity commands, calculated using PD controllers were sent to the UAV at frequency of 15Hz, ensuring real-time tracking, having in mind the slow changes in human movements, i.e. inputs to the system.

The proposed solution was tested and evaluated on 11 individuals, in laboratory and outdoor environments, size of $750m^2$. Complete solutions, testing videos and evaluations are available on GitHub[1]. Fig.4. presents tests conducted tracking across various body orientations and scenarios.

---

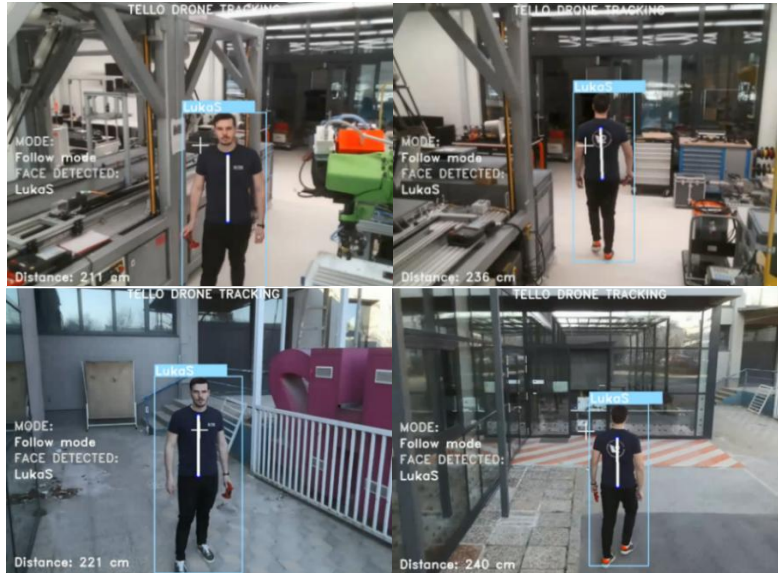

**Fig.4.** Proposed solution results, indoor and outdoor daily environment

## 5 Conclusion

This research presented a new approach of using UAVs for search and rescue, including people detection, face recognition and body tracking. The system leverages YOLOv11 for people detection, the dlib's library models for face recognition, and YOLOv11-pose for body keypoint estimation, offering a new approach using state-of-the-art algorithms, CNNs, and for search and rescue subsystems. For DJI Tello UAV, the dynamic system is identified, and the three individual PD controllers are fine-tuned for real-time human tracking, sending velocity commands to the drone at 15 Hz. The complete solution is developed using ROS2 environment. The proposed system improves upon the previous approaches that required the UAV to face the targeted person directly to the face, enabling flexible tracking based on person's movement. YOLOv11-body keypoints are utilized to successfully approximate the UAV's monocular camera-person distance. The system is accurate and suitable for real-time, real-world implementation.

Further work includes migration of the system to a more advanced UAV system with high resolution camera and stereo camera. High resolution camera will improve the YOLOv11 results, and stereo camera system can give more accurate proximity distance. Also, the next step is to mount all the computation, using Jetson NX Xavier, on the UAV to remove any wireless connection problems.

## 6 Acknowledgements

The authors would like to acknowledge the support of the project CRTA, funded by the ERDF fund and the ROBOCAMP, funded by the NPOO program.